\documentclass{article}

     \usepackage[preprint]{neurips_2019}

\usepackage[utf8]{inputenc} 
\usepackage[T1]{fontenc}    
\usepackage{url}            
\usepackage{booktabs}       
\usepackage{amsfonts}       
\usepackage{nicefrac}       
\usepackage{microtype}      

\usepackage{graphicx}
\usepackage{amsmath}
\usepackage{amssymb}
\usepackage{xcolor, colortbl}
\usepackage{verbatim}
\usepackage{tabularx}
\usepackage{pifont} 
\usepackage{subcaption}
\usepackage{array}
\usepackage[inline]{enumitem}
\usepackage[export]{adjustbox}

\usepackage[pagebackref,breaklinks,colorlinks]{hyperref}
\usepackage[capitalize]{cleveref}
\crefname{section}{Sec.}{Secs.}
\Crefname{section}{Section}{Sections}
\Crefname{table}{Table}{Tables}
\crefname{table}{Tab.}{Tabs.}
\setcitestyle{square}
\setcitestyle{citesep={,}}

\title{Midterm Report: Controllable Garment Transfer}

\author{
  Jooeun Son \\
  \texttt{jooeuns@andrew.cmu.edu} \\
  \And
  Tomas Cabezon Pedroso \\
  \texttt{tcabezon@andrew.cmu.edu} \\
  \AND
  Carolene Siga\\
  \texttt{csiga@andrew.cmu.edu} \\
  \And
  Jinsung Lee \\
  \texttt{jinsungl@andrew.cmu.edu} \\
}

\begin{document}
\maketitle

\begin{abstract}
Image-based garment transfer replaces the garment on the target human with the desired garment; this enables users to virtually view themselves in the desired garment. To this end, many approaches have been proposed using the generative model and have shown promising results. However, most fail to provide the user with on the fly garment modification functionality. We aim to add this customizable option of "garment tweaking" to our model to control garment attributes, such as sleeve length, waist width, and garment texture.
\end{abstract}

\section{Introduction}

Garment Transfer is a field of computer graphics concerning the photo-realistic transfer of a garment from a reference image onto a human. With a wide range of applications, some works focus on transferring garment images from catalogues, whereas others use more realistic images to extract garments. Research today mainly focuses on the transfer of the garment images only despite the possibility of combining garment modification tasks to increase user functionality. Although some systems include this functionality, it is primarily for texture transfers; in this case, the user needs to input another reference image and specify the modifications. In this work, we implement a virtual try-on that enables users to control attributes of the garment after the initial output. We call this functionality ‘garment tweaking’, providing a personalized try-on experience to the user. 

When we try on clothes, it is common to wonder how small changes in the garment will influence the fit. Modifying a garment to have longer sleeves, a wider waist, a more open neck, or even a different colour are some of the ‘tweaks’ we aim to include. The task is composed of two stages: garment transfer and garment tweaking. Our pipeline is sequential and is as follows —the user inputs a target garment and reference-human image; the system creates an image post the garment transfer; the user can further modify the garment using our garment tweaking functionality post which the system creates another image with the required modifications.

\section{Related Work}

The garment Transfer task was first proposed in \cite{jetchev2017conditional} and has recently gained interest due to its applicability in the merchandise industry. Approaches vary to many extents: some work focuses on cataloguing garment images, whereas others use in-the-world images to extract garments. There are two main branches in garment transfer attempts: 3D-model based approaches \cite{magnenat20113d, guan2012drape, patel2020tailornet, pons2017clothcap, sekine2014virtual} and 2D-model based approaches \cite{raj2018swapnet, han2018viton, yang2021ct, yang2020towards, raffiee2021garmentgan}. 3D-model based attempts can predict the garment on the target person in a 3d virtual space and accurately address the occlusion issues of the task but require 3D measurement data. On the other hand, 2D-image based attempts have access to far more data online and are computationally efficient, amplifying their applicability. Accordingly, we further dive into 2D-image based research in this section as we work on 2D images.

A successful garment transfer network must be robust to arbitrary human poses and occlusion while preserving the complex features of the garment and pose of the reference human. Earlier works maintain sub-networks to tackle each sub-task to cope with the complexity. In SwapNet~\cite{raj2018swapnet}, the task comprises two stages, where the first utilizes two encoders to extract pose and clothing representations from the input images individually. The two are fused to create the final result in the latter stage. VITON~\cite{han2018viton} also utilizes a similar structure; the main difference is that the two encoder networks are sequentially connected so that one encoder network refines the coarse output of the other encoder-decoder network.

Recent works suggest more sophisticated designs, with separate modules for warping, layout prediction and fusion. CT-Net~\cite{yang2021ct} estimated two complementary warpings(DF-guided dense warping and TPS warping) to deal with the different levels of geometric changes in the transfer process. In ACGPN~\cite{yang2020towards}, they progressively generate the mask of exposed body parts using the semantic segmentation of the input images. GarmentGAN~\cite{raffiee2021garmentgan} uses two streams of GAN networks: one to generate a new shape map given person representations, semantic segmentation of a reference person, and target cloth, and the other to generate the visually transformed output using the previously formed map in the subsequent decoding process. The most related work is DiOR~\cite{cui2021dressing}, which incorporates the way people dress in real life. DiOR extracts the target pose and each garment feature from the target image and sequentially applies each feature to the source person by utilizing an encoder-decoder module that enables a sequential try-on process and provides different results based on the order of trial.

\section{Methodology}
\subsection{Baseline Selection}

The chosen baseline model is \textit{Dressing in Order} (DiOR)~\cite{cui2021dressing}. We justify our thought process in this section.

\noindent \textbf{Baseline Criteria}
There were several criteria that we set up to select our baseline:
\begin{enumerate}
  \item Are source code and dataset publicly accessible?
  \item If so, is the source code in Python?
  \item If so, is it based on Pytorch library?
\end{enumerate}
We explored a myriad of studies~\cite{han2018viton, raj2018swapnet, yang2021ct, raffiee2021garmentgan, choi2021viton, yang2020towards, jong2019short}, and these did not qualify the criteria mentioned above. The datasets that were open to the public were VITON~\cite{han2018viton} and DeepFashion~\cite{liu2016deepfashion}. Among the works that have released the source code, only SwapNet~\cite{raj2018swapnet} and DiOR~\cite{ cui2021dressing} used Python and Pytorch as their major programming language and deep learning package.

\noindent \textbf{Model attributes}
\begin{enumerate}
  \item Is the work \textbf{recent}ly published, is it still relevant?
  \item Are the training time and inference time not overwhelmingly long?
  \item Is there room for improvement that we can fix?
  \item Is it compatible with our idea of garment modification which would add controllability to the baseline? 
\end{enumerate}

DiOR is much more recent than SwapNet: DiOR was published in 2021, while SwapNet was published in 2018. A 3-year gap exists between these two; being a more recent model, we expected DiOR to have better performance. Surprisingly, when compared based on the SSIM metric, SwapNet showed superior performance. DiOR, on the other hand, maintains an upper hand while handling various attributes in a customizable way and demonstrates robustness to various poses of the target garment. The training time and inference time of both models were not too long. SwapNet requires two stages of training, and training both stream takes up to 40 hours. DiOR takes up a bit longer: requiring a time of 48 hours for the training process. Note: the 1 RTX 3080ti GPU was used to compute training/inference time. Both papers listed their limitations, and we also ran the inference code to see the failure cases. DiOR displayed concerns when transferring garments, while SwapNet had issues transferring pose. Besides implementing tweak-ability inside the model, we observed room for model quality improvement in both experiments.

The most significant concern was the ability of the baseline model to accommodate our garment tweaking module. SwapNet extracts the clothing features implicitly in the first stage and fuses with the target image's pose representation. Thus, when placing our tweaking module, it would need to be placed inside the warping stage before the clothing representation is fused with the target pose. However, this is done simply by the encoder module and the features we want to control need to be decoupled from the latent space of extracted representation. However, the architecture of DiOR divides attributes of the target image and progressively applies each feature sequentially, which makes our feature controller more manageable.

\subsection{Model Description}

\begin{figure}[ht]
    \centering
    \includegraphics[width=1.0\textwidth]{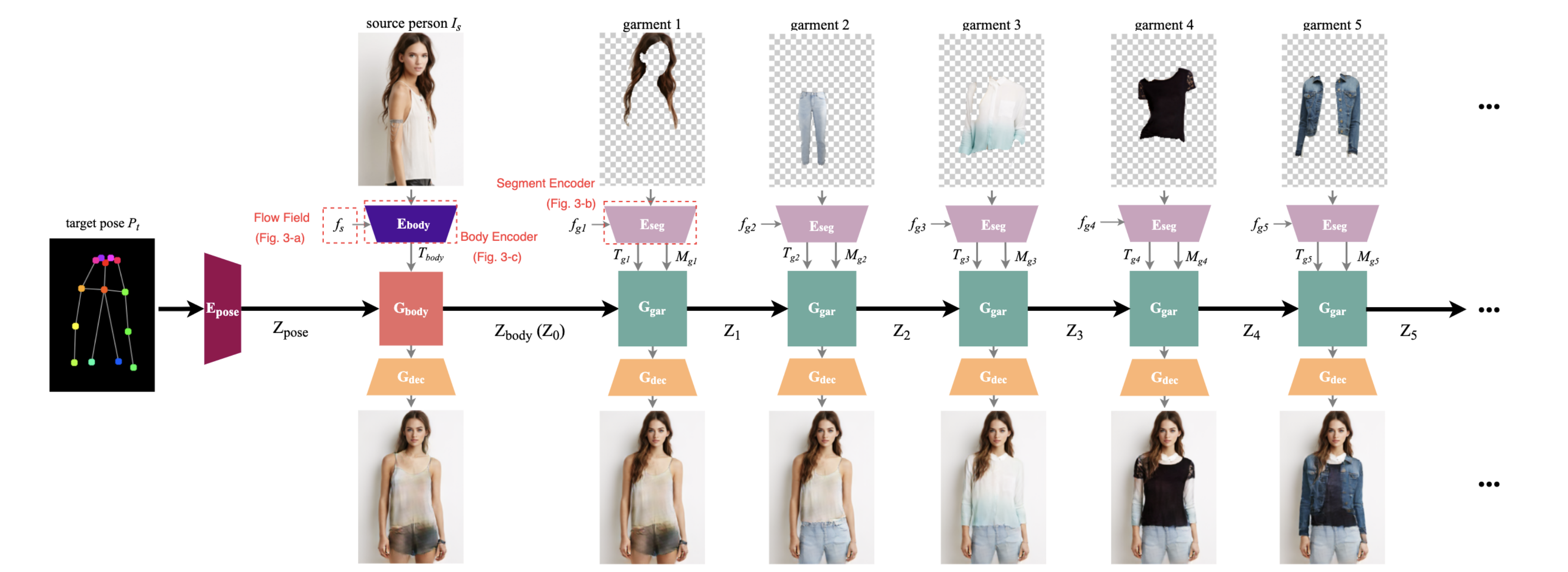} 
    \caption{DiOR generation pipeline.}
    \label{fig:pipeline}
\end{figure}


Consequently, the model selected for this work is the recently published framework DiOR. \cite{Cui_2021_ICCV}. This work is inspired by the previous work of 2020 Attribute-decomposed GAN (ADGAN) \cite{men2020controllable}. The main difference between this approach and the previous one is that DiOR encodes the garment's shape and texture separately in 2D instead of 1D like ADGAN. As the results show, this is a better approach, both in terms of performance and the flexibility incorporated in the model, allowing these elements to be edited separately. In the DiOR generation pipeline, every person is represented by the tuple \textit{(pose, body, \{garments\})}. As it can be seen in Fig. \ref{fig:pipeline}, the process is separated into several steps that are combined to achieve the final output image. 

\subsubsection{Person Representation}

The initial step is the pose generation which is done by the Pose Encoder $E_{pose}$; it encodes the target pose as $Z_{pose}$. To encode this target pose $P_{t}$, it is represented as 18 keypoint heatmaps defined by OpenPose \cite{cao2017realtime}. 

\begin{figure}[ht]
    \centering
    \includegraphics[width=0.7\textwidth]{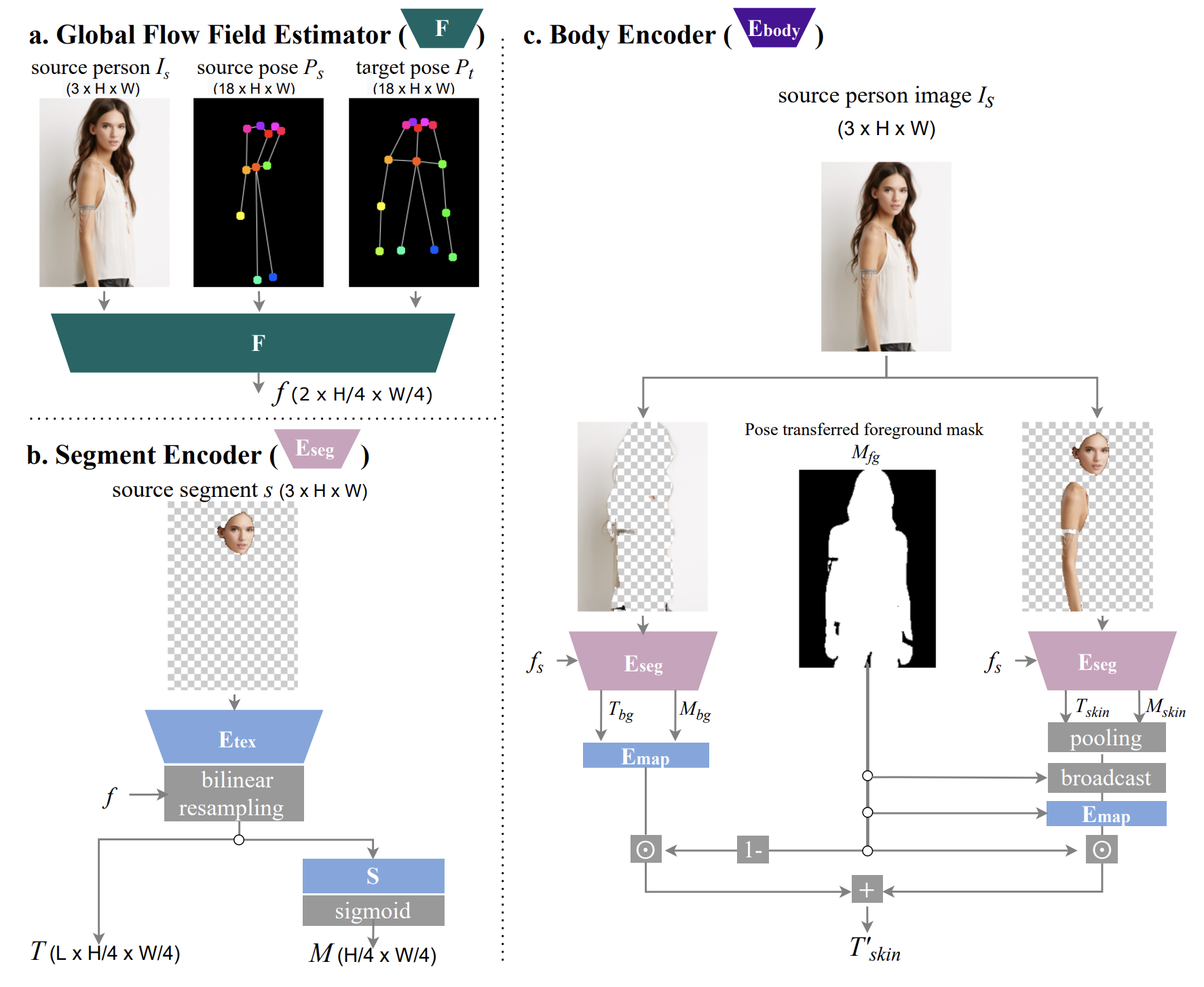} 
    \caption{DiOR system details.}
    \label{fig:details}
\end{figure}

\textbf{Garment representation}. To acquire the garment feature, a garment $g_k$ worn by a person in an image $I_{gk}\in \mathbb{R} ^{3 \times H \times W}$ is extracted using the off-the-shelf human parser to generate the masked garment segment $s_{g_k}$. Because the target pose $P$ and the pose in the image $P_{g_k}$ are not same, a flow field $f_{g_k}$ needs to infer to align the pose $P$ with the garment segment $s_{g_k}$. This is done by using the Global Field Estimator \textbf{F} from GFLA. The masked garment segment will be encoded by the following segment encoder module $E_{seg}$. This starts with a texture encoder $E_{tex}$ whose output is warped by the previously calculated flow field $f_{g_k}$ using bilinear interpolation, which results in the texture feature map $T_{g_k}$. At the same time, a soft shape mask of the garment segment is computed $M_{g_k}=S(T_{g_k})$, where $S$ is a segment consisting of three convolutional layers. Both $T_{g_k}$ and $M_{g_k}$ are the outputs of the segment encoder $E_{seg}$. The garment feature extraction process is illustrated in Fig. \ref{fig:details} \textbf{a} and \textbf{b}.

\textbf{Body representation}. This part encodes the body of a source person in image $I_{s}\in \mathbb{R} ^{3\times H\times W}$. The same human segmentation module that is occupied in the previous part extracts the background and skin masks $s_{bg}$ and $s_{skin}$, respectively. The latter one consists of the arms, legs and face. These masks are encoded by the previously described segment encoder module $E_{seg}$, which outputs  $(T_{bg} , M_{bg} )$ and $(T_{skin} , M_{skin })$. The body representation extractor module is described in Fig. \ref{fig:details} \textbf{c}.

The body feature map may not span the entire body depending on the garment the person is wearing in the image; the entire body region is computed by computing the mean body vector $b$ over the ROI defined by $M_{skin}$ to avoid this. This vector $b$ is broadcasted to the pose-transferred region $M_{fg}$, which is the union of all the pose transferred foreground masks. It is then mapped to match the correct dimension by $E_{map}$ and finally combined with the mapped background feature $ T'_{bg} = E_{map} (T_{bg} , M_{bg} )$ to get the body texture map $T'_{body}$:

\begin{equation}[ht]
    T'_{body} = M_{fg}\odot E_{map}(M_{fg}\otimes b,M_{fg})+(1-M_{fg})\odot T'_{bg}
    \label{eq:T'body}
\end{equation}

where $\otimes$ and $\odot$, represent broadcasting and element-wise operations, respectively.

\subsubsection{Generation pipeline}

Fig. \ref{fig:pipeline} illustrates the pipeline of encoding the pose $P$, generating the body from $T'_{body}$, eq. \ref{eq:T'body}, and adding the sequence of garments, each encoded by $(T_{g_{i}} , M_{g_{i}} )$.

\textbf{Pose and skin generation}. The pose $P$ is encoded by the pose encoder $E_{pose}$ that is composed of three Convolution-InstanceNorm-LeakyReLU layers. Therefore, the encoded hidden pose map $Z_{pose} \in \mathbb{R} ^{L\times H/4\times W/4}$ is yielded, where L is the latent channel size.

Then, using the body generator $G_{body}$ and given $Z_{pose}$ and the texture map $T'_{body}$, we generate the hidden body map $Z_{body}$. This $G_{body}$ generator is implemented by two ADGAN \cite{men2020controllable} style blocks. As previously mentioned, the main difference between DiOR and the previous ADGAN work is that each garment is encoded in shape and texture separately in 2D. Therefore, the adaptative instance normalization used in ADGAN is replaced by SPADE \cite{Park_2019_CVPR}. Furthermore, the previously described $E_{map}$ is used to convert the style input in the desired dimensions.

\textbf{Recurrent garment generation}. Once we have enconded $Z_{body}$, now called $Z_0$, we will sequentially add the garments. We denote $T'_{g_k}$ the the texture map and $M_{g_k}$ the soft shape mask of the \textit{k}-th garment. Using the garment generator $G_{gar}$ and putting these two together with the previous state $Z_{k-1}$, we produce the next state $Z_{k}$ as:

\begin{equation}[ht]
    Z_{k}=\boldsymbol{\Phi}\left(Z_{k-1}, T_{g_{k}}^{\prime}\right) \odot M_{g_{k}}+Z_{k-1} \odot\left(1-M_{g_{k}}\right)
    \label{eq:Z_k}
\end{equation}

where $\boldsymbol{\Phi}$ is a conditional generation module with the same structure as $G_{body}$ above.

After all the garments are added, the final image $I_{gen}$ is generated from the final hidden feature map $Z_K$ using the $G_{dec}$ decoder, implemented in the same way as in the ADGAN \cite{men2020controllable}, consisting of residual blocks, upsampling and convolutional layers followed by layer normalization and ReLU.

\subsubsection{Training}
The training process of the DiOR is similar to the training of the ADGAN \cite{men2020controllable}. Given a person image $I_s$ in a source pose $P_s$, we will generate that person in a target pose $P_t$; this will be a supervised task as long as we have the reference image $I_t$ of the same person in the target pose.

The authors of DiOR \cite{ cui2021dressing} saw that training the whole process was better than training the pose transfer and garment transfer separately. 

To train the model, they used the loss eq. \ref{eq:training loss} shown below. These terms are the same terms from GFLA. The $L_{content}$ is the combination of the L1 loss, perceptual loss, and style loss of the generated and real target pairs. The second term, $L_{geo}$, is the geometric loss of the combination of the correctness and regularization losses for the predicted flow fields. The $L_{GAN}$  are the losses of the two discriminators used for pose and segmentation. These discriminators have the same architecture as GFLA. The last term $L_{seg}$ is used to ensure that the shape masks have the correct shape. This is a pixel-level cross-entropy loss between the shape mask $M_g$, and the extracted segmentation of the garment using the parser. The final loss equation is given by: 

\begin{equation}[ht]
    L=L_{content}+L_{geo}+\lambda_{GAN}L_{GAN}+\lambda_{seg}L_{seg} 
    \label{eq:training loss}
\end{equation}

\subsection{Evaluation Metrics}
In the garment transfer literature, there are commonly used metrics: SSIM~\cite{wang2004image}, FID~\cite{heusel2017gans}, and LPIPS~\cite{zhang2018unreasonable}. All metrics are devised to measure the implicit similarity between generated image and the real reference image. The authors of the original paper also proposed a new metric \textbf{sIOU}, which is the average of the IoU of the segmentation masks produced by the human segmenter for real and generated images. In our future progress, we will also measure the efficiency of our proposed module through the listed metrics.

\section{Progress}

Our baseline requires two preprocessing steps before being able to create V-TON results: pose estimation and segmentation. We have done qualitative tests of both these tools on custom images and concluded that they were both capable of performing each task. In addition to this, we tested out two main functionalities claimed in the paper and quantitative evaluation for the task of pose transfer.
\subsection{Preprocessing Step}

The network requires three things as input: the source image, target pose, and segmented garment images. The following steps are required to preprocess the target pose and garment images to be used as input to the network.

\textbf{Pose Estimation}. 
To create the target pose, this is needed to extract the human key points from a human image. The baseline refers to OpenPose \cite{cao2017realtime} for this task, whose path we followed.

\textbf{Human Parser}. In addition to that, the garments need to be segmented prior to being input to the system and the exact segment index that each turn of the generation process focuses on needs to be explicitly stated.

\subsection{Baseline}
The following sections describe the results obtained from the exploration we have done on the baseline. To validate the result of the paper, find the room for improvement, and get familiar with the code structures of the baseline, both qualitative and quantitative evaluations are conducted.
\subsubsection{Qualitative Evaluation}
We have tested mainly two ideas claimed in the baseline.

\textbf{Pose Transfer}. 
The first is pose transfer, which is the backbone task of the baseline. This task does not involve any garment transferring but aims to transform the source image's human in the target pose.

\begin{figure*}[ht]
     \centering
     \begin{minipage}[b]{0.40\textwidth}
         \centering
         \includegraphics[width=\textwidth, valign=t]{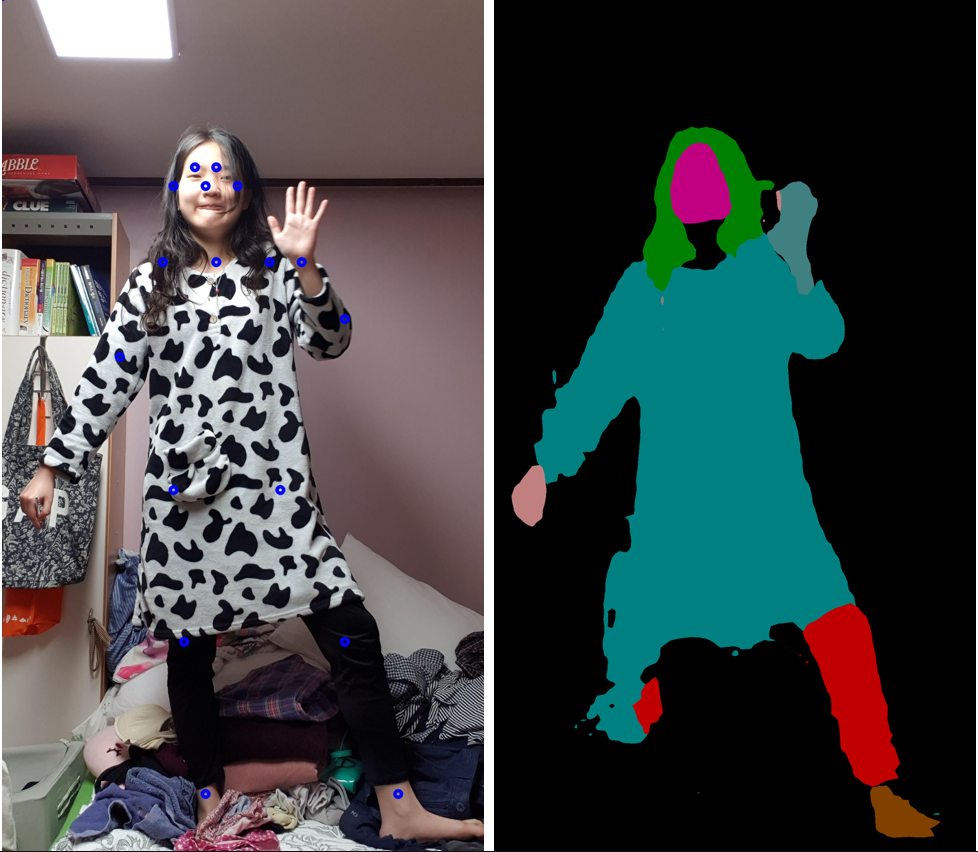}
         \caption{Pose estimation and Parsing results tested with custom data. 
         \vspace{44pt}}
         \label{fig:PES}
     \end{minipage} \hspace{1mm}
     \begin{minipage}[b]{0.57\textwidth}
         \centering
         \includegraphics[width=\textwidth, valign=t]{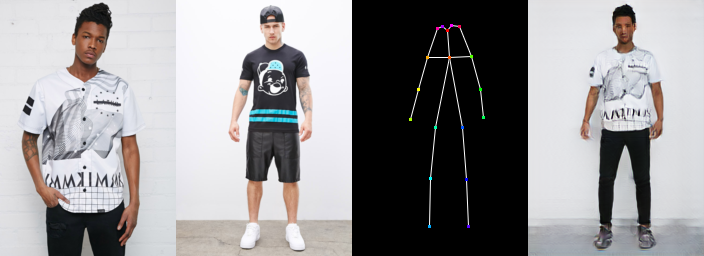}
         \includegraphics[width=\textwidth, valign=t]{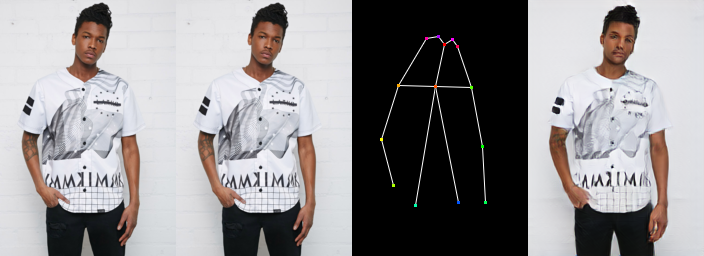}
         \caption{From the left: source image, target image, pose extracted from target image, pose transfer results. \textbf{Top} Source image pose transferred to that of the target image. \textbf{Bottom} Results when providing target image identical to the source image. Results show failure to preserve identity (face).}
         \label{fig:src-tgt}
     \end{minipage}
     \vspace{-2mm}
     \vspace{-2mm}
\end{figure*}

Fig~\ref{fig:PES} shows the pose transfer results. We found that although the network can transfer poses at a high level, it fails to preserve the detail and identity of the source image. For example, the bottom image of Fig~\ref{fig:src-tgt} shows the pose transfer results when using the pose extracted from the source image. Although the intuitive result would be to return the same image as the original, it messes up the key parts of the original image, such as the face.

\begin{figure}[ht]
\includegraphics[width=\textwidth]{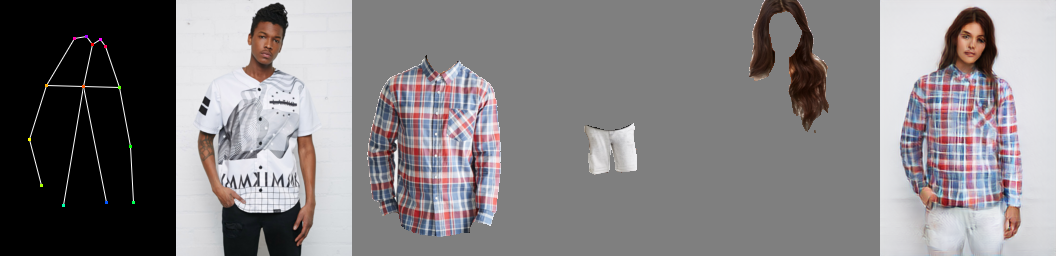}
\includegraphics[width=\textwidth]{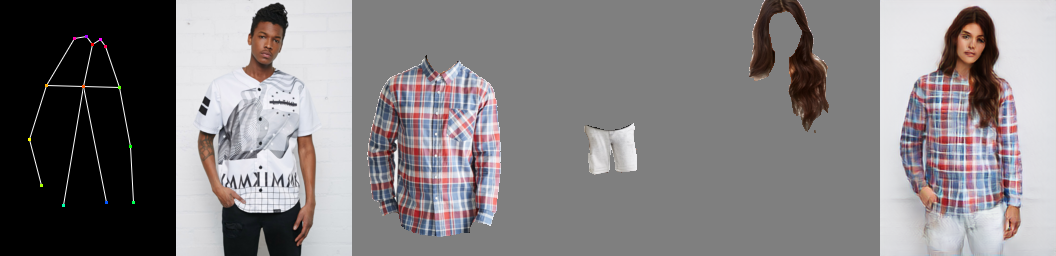}
\includegraphics[width=\textwidth]{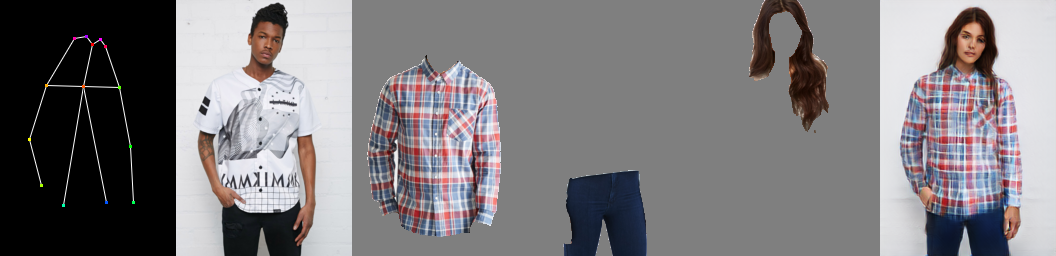}
\includegraphics[width=\textwidth]{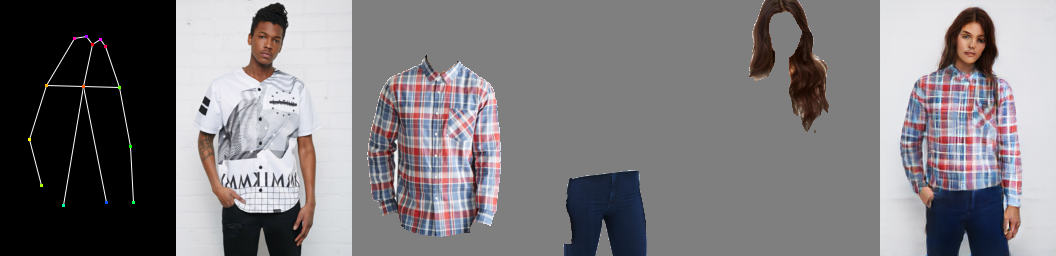}
\caption{Order variation results. From the left: extracted pose from the source, source image, segmented shirt, pants, and hair (the "garments"), synthesis results. From the top, the input order is hair-shirt-pants, pants-shirt-hair, hair-pants-shirt, and hair-shirt-pants. The \textbf{first two} figures illustrate the effect of changing the order of the hair. The \textbf{last two} figures illustrate the 'tuck-in' effect.}
\label{fig:ord-var}
\centering
\end{figure}

\textbf{Order Variation}. 
The main advantage of our baseline lies in its flexibility. It not only allows multiple garments as input, but the authors also claim that changing the order of the garments will create different outputs. In Fig~\ref{fig:ord-var}, we mainly tested out different orderings of the hair, shirt, and bottom and found that the network, in general, can create different layering effects according to the given garment order.

\subsubsection{Quantitative Evaluation}
In the original paper, the authors provide the quantitative performance of their model for the task of pose transfer in terms of SSIM (Structural Similarity Index Measure). SSIM is measured between generated and the reference image and is used as an evaluation metric in the garment transfer task. It is stated that the model has achieved an SSIM of 0.806 when tested on the test dataset. We implemented the code using a subset of images from the test data and acquired the SSIM of 0.80444, which validates the result mentioned in the paper.

\section{Future Steps}
\subsection{Identifying Limitations}

Although our baseline model is competent rare poses and garment shapes are not always appropriately transferred- the shading, texture warping, and garment detail preservation are still not entirely realistic. Implementing more advanced warping and higher-resolution training and generation could improve results. Additionally, in some test cases, we noticed that the user's facial features were morphing with that of the model of the reference image (when models are wearing the garments). We will be working towards reducing this.

\subsection{Proposed Extensions}

We will be incorporating the "garment tweaking" application into our baseline model. The first "tweak" that we will be looking into is the length of the sleeve and bottoms. We look into the semantic garment editing by interpreting the latent semantics learned by GANs. We adapt this idea from existing work\cite{SemanticFaceEditing} that focuses on analysing the latent space of GANs for semantic face editing.

{\small
\bibliographystyle{plainnat}
\clearpage
\bibliography{main.bib}
}
\end{document}